# TRUTH AS UTILITY:
# A CONCEPTUAL SYNTHESIS

**Enrique H. Ruspini**
Artificial Intelligence Center
SRI International
Menlo Park, CA 94025

## Abstract

This paper introduces conceptual relations that synthesize utilitarian and logical concepts, extending the logics of preference of Rescher.

We define first, in the context of a possible-worlds model, constraint-dependent measures that quantify the relative quality of alternative solutions of reasoning problems or the relative desirability of various policies in control, decision, and planning problems.

We show that these measures may be interpreted as truth values in a multivalued logic and propose mechanisms for the representation of complex constraints as combinations of simpler restrictions. These extended logical operations permit also the combination and aggregation of goal-specific quality measures into global measures of utility. We identify also relations that represent differential preferences between alternative solutions and relate them to the previously defined desirability measures.

Extending conventional modal logic formulations, we introduce structures for the representation of ignorance about the utility of alternative solutions. Finally, we examine relations between these concepts and similarity-based semantic models of fuzzy logic.

## 1  Introduction

The ability of logic-based procedures to represent knowledge elements of rather diverse characteristics while identifying, by constructive proof, solutions of a wide variety of problems is the major reason for their appeal as the bases of a class of artificial intelligence methodologies.

Aristotle, who established logic as the discipline concerned with the relations between the truth-values of propositions, was also, as noted by Rescher [Rescher 67], the first student of the notion of preferability ("the worthier of choice"), also a major element of any problem-solving approach.

A large amount of interest has been recently expressed in the artificial intelligence community about the role of the concept of utility in the solution of various problems. Hobbs et al. [Hobbs 90] have, for example, recently proposed measures of cost to assess the quality of interpretations of linguistic utterances. The work of Russell and Wefald [Russell 89] also exemplifies recent interest on the applicability of decision-theoretic principles to the control of reasoning processes. Measures of preference have also been used in intelligent decision planners to control the relaxation of "elastic" constraints [Fox 90].

The appeal that utilitarian notions and structures have as an important element of the problem-solving process lies on their ability to add to the classical logicist concept of rationality—the sound derivation of conclusions from premises— pragmatic principles of rational behavior that aid in the evaluation of the desirability and utility of both assumptions and conclusions, often assisting to determine the relative importance of various constraints. Furthermore, at a metareasoning level, utilitarian considerations are important elements to guide reasoning processes along lines that are likelier to produce the truth values of target hypotheses.

In this paper, we advance a view of truth as utility that follows naturally from considerations about the relative desirability of alternative solutions and that effectively integrates both notions by means of multivalued logic approaches. Our approach is influenced by the "logics of preference" proposed by Rescher [Rescher 67] where the truth-value (usually measured in a $[0, 1]$ scale) of a constraining proposition $p$ represents the desirability of $p$ coming about, or, in other words, the degree by which $p$ is a "good thing." While being close in spirit to the same multivalued-logic ideas, the approach followed by this paper differs from that of Rescher in a number of substantial re-gards.

First, we seek to develop a model that associates a utility measure, measuring the relative goodness—from a particular perspective—of the solutions of a problem "all other things being equal," to each constraint that defines the problem. Our measures quantify preference between alternatives (e.g., being on this or the other side of the street) from the limited perspective of a single constraint (e.g., we want to hire a taxi). This formulation represents a substantial departure from Rescher's formalism where utilities are simply functions of propositions that measure their "global" relative desirability, regardless of context. This global measure, in Rescher's approach, is given by an average of context-specific desirability values: an assumption that leads to the narrow conclusion that such measures must have the properties of probability distributions.

Second, by introduction of modalities, we enhance the value of multivalued-logic schemes, generalizing both our previous semantic models of fuzzy logic [Ruspini 91] and providing a practical way to represent ignorance about the potential utility of certain choices.

Finally, we provide bases for the rational combination of multiple, goal-specific, measures of utility into a global preference relation that represents their relative importance.

Beyond our objective of expanding and exploring the notions proposed by Rescher, we seek to establish and study formal bases for utility-oriented approaches to problem-solving that depart from classical "optimization" methods that maximize a prespecified measure of performance subject to "hard" constraints. Our formalism, which is also inspired by recent studies on operations research and the foundations of utility theory [Brachinger 90], regards every constraint on the solution of a problem as a source of differential preference relations between alternatives, which are then traded off by a metareasoner capable of *explaining* its solution rationale; a task beyond the ability of current optimization techniques. Furthermore, each such constraint has a "target value," i.e., a quantitative characterization of the "ideal" state of affairs from the single perspective of that restrictive statement.

Our model is based on the mapping of each potential solution of a problem to a set of numbers, each representing the desirability of that alternative from the viewpoint of a different constraint. This perspective on the reasoning problem leads to a uniform characterization of the role that constraints of diverse type have on the determination of the suitability of solutions.

There is no need, for example, to associate two measures, one representing utility and the other representing cost, to each goal or constraint: a most convenient methodological property that is easier to appreciate by noticing that formalisms requiring such one-on-one associations fail to capture complex relations between constraints (i.e., expenditure of some resource contributes to the attainment of multiple goals while attainment of any goal entails use of several resources). Constraints that require attainment of some goal (e.g., "buy nice presents for Jim and Joe") and constraints that limit resource expenditure (e.g., "do not spend too much") are both sources of the same type of measures, which evaluate the desirability of different state of affairs (i.e., "buying presents" is a source of preference measures, and "not spending too much" is another such source). Furthermore, every constraint, including those describing system behavior (e.g., the laws of physics), is the source of utility measures (e.g., as functions of the costs associated with simplifying assumptions that are usually not met by any real system, such as "the container is filled with a perfect gas").

Our ideas also owe much to previous research, which established the close relationships between generalized rankings, operations research, and fuzzy logic [Bellman 80] and similarity-based semantics for that approximate logic [Ruspini 91]. In addition, we have been particularly motivated by the need to provide practical bases for the generation of similarity-measures— the conceptual foundations of analogical reasoning methods— on the basis of the pragmatic rationale for the differentiation between alternative solutions, i.e., two scenarios resemble each other if they are equally preferrable in every relevant regard.

In closing this introduction, it is very important to remark that our synthesis of utilitarian and logical concepts is not the result of a trivial confusion of what is true and what is convenient. We are simply stating that the truth value, measured in a multivalued scale, of propositions of the form:

"The possible world $s$ is an acceptable solution,"

may be interpreted as the relative degree of preference given to $s$ from a specific viewpoint.

## 2   Possible Worlds and Desirabilities

Our formalism is based on the notion of *possible world*, which will only be given a brief, informal, characterization in this paper. Basically, a possible world, is any conceivable scenario, situation, or behavior that may be used to describe the state of a real-world system. Each such situation is modeled by a function, called a *valuation*, that assigns a conventional truth-value (i.e., either true or false) to every descriptive statement, or proposition, about the system. While we will require that such truth-assignment functions be consistent with the rules of logic, we will not place at this time any other restrictions on the nature of such assignments. Thus, possible worlds may corre-spond to impossible physical situations, such as the state of idealized systems (e.g., "perfect gases").

The importance of the concept of of possible world, from a reasoning viewpoint, lies on its usefulness to model solutions of reasoning problems as subsets of possible worlds that satisfy certain constraints (i.e., observations, behavioral knowledge) that are usually called "evidence." From such a purely logical viewpoint, a reasoning problem consists of the determination of the set of possible worlds that comply with prescribed constraints. In a classical reasoning problem, the properties of interest are related to the relations of inclusion that hold between the set of possible worlds that satisfy the evidence and sets of possible worlds where some proposition of interest, or *hypothesis* is true. In a number of important cases where such determination is impossible, or *approximate reasoning* problems, applicable approaches, described as approximate reasoning methods, seek to determine certain properties of the set of possible worlds that are consistent with the evidence [Ruspini 90].

Regardless of the nature of our problem, however, we may describe the object of reasoning as being that of assessing if a possible world is worthy of being called a solution. The use of the qualifier "worthy" in the above sentence is not accidental. From a pragmatic point of view, we may say that our aim is to find useful solutions, where "usefulness" is determined by compliance with problem conditions. Given more representational freedom, we may decide to rank (e.g., using some quantitative measures) such usefulness and to speak of the "quality" of a solution.

This notion of utility as cognate of truth is rather easy to understand in connection with engineering problems, where the quality of the solution reflects the ability of a device or system to perform adequately. In general, however, we may regard any problem-solving process as the identification of models (i.e., sets of possible worlds) that meet stated constraints—including those imposed by measurements and observations—to some acceptable degree.

In summary, we may say that we seek solutions with a number of properties, corresponding to compliance with constraints, and that we are willing to pay to different extents to see that those properties come about. At this point, however, it is important to make a few points that are essential to equate measures of desirability with measures of truth.

First, we may note that, in a conventional Boolean context, every proposition $p$ is equivalent to a measure of the solution quality, as measured solely from such a viewpoint (i.e., "other things being equal"). In our formalism, we may say that we have a function $\mathbf{D}_p$, called a *desirability* measure, that assigns a value of 1 to every possible world where $p$ is true (denoted $w \vdash p$) and a value of 0 to every possible world where it is false.

In this case, this *characteristic function*, in categorical fashion, what is acceptable and what is not. It is often the case, however, that the desirability of a solution is a matter of degree. In such cases, a most natural generalization of the notion of characteristic function will be of value to model such graded preferences: the concept of *fuzzy set* [Zadeh 65].

Second, we must recognize that it will not be difficult to find problems where situation-dependent wishes and desires may not be easily expressed as a single number that quantifies preference. Furthermore, if we aim to associate the truth of any proposition $p$, not necessarily specified as a problem constraint, with some utilitarian counterpart, we must have some mechanism to indicate that compliance with such proposition is irrelevant to the solution of the problem. The simplest mechanism to represent such irrelevance is to identify a family of *possible* desirability measures to be associated with $p$, which is easily done by identification of upper and lower bounds for $\mathbf{D}_p$, thus generalizing the notions of possibility and necessity of modal logic.

Finally, we must remark that our functions, which are constraint-dependent, represent relative adequacy of alternative solutions with respect to a set of ideal solutions that attain or exceed some associated "target value." Such ideal solutions are given a relative measure of adequacy that is equal to 1.

It should be clear, however, that the "absolute" usefulness associated with the satisfaction of a particular goal should be given by some "utility" function (defined along some suitable scale of measurement) that, loosely speaking, represents the overall utility associated with its achievement (e.g. if utility is measured in a monetary scale, we may say that the utility of $p$ coming about is $100). To limit the scope of this paper, however, we shall confine ourselves to the discussion of issues germane to desirability measures, avoiding questions related to the absolute utility associated with their achievement.

## 3    Desirability and Preference

As informally introduced above, desirability measures quantify the relative value of different solutions from the viewpoint of a single constraint or goal. Although that discussion was confined to "nonelastic" goals, corresponding to subsets of possible worlds, the most interesting applications of utilitarian concepts involve measures ranging over a continuous scale.

### 3.1    Desirability Measures

The simplest way to formalize the notion of adequacy of a solution is that of a measure that assigns a value of relative desirability to any conceivable solution, as determined solely from the viewpoint of a single, specific goal. In its simplest form, a solution corresponds to some fully specified description (i.e., a possible world) and it makes sense to start our formalism with a real function defined over possible worlds ranging from 0, representing total inadequacy, to 1, representing total compliance or satisfaction with the goal.

**Definition:** A *desirability measure* is a function $\mathbf{D} : \mathcal{U} \mapsto [0, 1]$, i.e., a fuzzy set in the universe $\mathcal{U}$ of possible worlds.

The concept of desirability measure is a natural extension of the notion of "hard" or "crisp" constraint. The values $\mathbf{D}(w)$ may also be thought of as the truth-values of the proposition "The solution $w$ is satisfactory." The conceptual synthesis between desirability measures and propositional truth that this explication implies leads to the extension of classical propositional algebra, along well known lines, into a multivalued methodology for their rational combination and aggregation.

Before discussing such methods, however, it is important to remark, once again, that this conceptual unification should not be interpreted as an attempt to reduce issues of factual truth to matters of subjective convenience. Our view of truth as utility stems from the same epistemological principles that led Peter Medawar to describe science as "the art of the solvable." We aim to qualify solution adequacy by measuring the extent to which potential answers fit factual reality and the constraints of the problem. Furthermore, we must stress that desirability measures quantify relative preferences between solutions, from a limited perspective, rather than the overall desirability of a proposition to come about or its importance among various problem constraints.

## 3.2 Relations between Desirability Measures

It would be rather odd if we were to say that a particular solution $w$ is desirable from the viewpoint of a constraint $p$, and that it is also desirable from the limited perspective of another constraint $q$, but that it is rather undesirable from the viewpoint of the conjunction $p \wedge q$. Rational considerations [Trillas 85] show that desirability measures ranking possible worlds by their ability to satisfy the conjunction of two constraints, themselves expressed by means of the desirability functions $\mathbf{D}$ and $\mathbf{D}'$, are related to such measures by the relation

$$(\mathbf{D} \wedge \mathbf{D}')(w) = \mathbf{D}(w) \circledast \mathbf{D}'(w), \quad w \text{ in } \mathcal{U},$$

where the function $\circledast$ is a *triangular norm*.

Similarly, desirability measures quantifying the degree by which solutions meet the disjunction of two restrictions can be seen to be given by

$$(\mathbf{D} \vee \mathbf{D}')(w) = \mathbf{D}(w) \oplus \mathbf{D}'(w), \quad w \text{ in } \mathcal{U},$$

where $\oplus$ is a *triangular conorm*.

Since it is reasonable to ask that the desirability of the conjunction and of the disjunction of two goals should not have an abrupt change when the desirabilities of the arguments are subject to slight variation, we would also require that $\circledast$ and $\oplus$ be continuous functions of both parameters.

Furthermore, desirability measures that rank possible solutions by the degree by which they do *not* meet some constraint (expressed by a desirability $\mathbf{D}$) are given by expressions of the form $\sim \mathbf{D}$, where $\sim$ is a *strong negation function*.

It can also be seen that desirability measures that rank possible worlds by the extent by which they satisfy a conditional constraint of the form $p \rightarrow q$, are related to the desirabilities $\mathbf{D}$ of $p$ and $\mathbf{D}'$ of $q$ by the relation: $(\mathbf{D} \rightarrow \mathbf{D}')(w) = \mathbf{D}'(w) \oslash \mathbf{D}(w)$, where $\oslash$ is the pseudoinverse of a triangular norm $\circledast$.

The discussion above simply restate results in fuzzy-set theory [Bellman 80,Dubois 84,Trillas 85,Valverde 85] that have been recast here in the context of a possible-worlds model to emphasize the parallelism that exists between desirability measures and generalized truth-values. Actual choice of particular T-norms, conorms, and negations depends on the semantics of the problem being considered, as a conjunction $\mathbf{D} \wedge \mathbf{D}'$ modeled using the minimum T-norm has considerably different properties than one being modeled using the product T-norm; being, in the former case, an assertion that minimum standards for $\mathbf{D}$ and $\mathbf{D}'$ must be met, while in the latter declaring that the degree of satisfaction of one goal is exchangeable with the degree of satisfaction of the other.

## 3.3 Preference Relations

The assignment of desirability values to diverse propositions is often made using comparative measures that assess the advantage that a particular solution $w$ has over a competing alternative $w'$ from the viewpoint of a specific constraint. The ability to define such comparative functions often simplifies the evaluation of the effect of contextual considerations (e.g., the desirability of being in another place, if we are thirsty, depends on how much water we have and how much water is in the other location).

We will formalize this notion by considering functions of the form $\rho(w|w')$ that map pairs of possible worlds to numbers between 0 and 1 so as to quantify the extent to which a possible world $w$ is preferred to another $w'$, from the viewpoint of a particular constraint. We may think of $\rho(w|w')$ as a measure of the amount of resources that we would be willing to spend to be in $w$ rather than in $w'$. It is easy to see that any definition for $\rho$ must comply with the following rational principles:

1. No resources should be spent to be in $w$ if we are

already in $w$.

2. If we are willing to spend resources to be in $w$ when we are in $w'$, then we should not spend any resources to be in $w'$ if we were in $w$

3. The amount that we would be willing to pay to be in $w$ when we are in $w''$ should be bound by above by a function of the amount that we would be willing to spend to be in $w$ if we were in $w'$ and of the amount that we would be willing to pay to be in $w'$ if we were in $w''$.

These principles are captured by the following

**Definition:** A function $\rho$ mapping pairs of possible worlds into numbers between 0 and 1 is called a $\oplus$-*preference* relation if and only if

1. $\rho(w|w) = 0$ for all $w$ in $\mathcal{U}$.

2. If $\rho(w|w') > 0$, then $\rho(w'|w) = 0$ for all $w$ and $w'$ in $\mathcal{U}$.

3. For any possible worlds $w$, $w'$ and $w''$ it is
$$\rho(w|w'') \leq \rho(w|w') \oplus \rho(w'|w'').$$

It is also easy to see that if $\rho$ has the semantics of a relation representing graded preference, then $\oplus$ should be a conorm.

### 3.4 Relations between Desirabilities and Preferences

The combination and aggregation of preference relatons is considerable more complex than that of desirability measures as, for example, the negation $\sim \rho$ of a preference relation $\rho$ is not itself a preference relation. In order to develop an aggregation methodology, it is necessary first to study the relations that exist between both types of utilitarian measures.

The derivation of a $\oplus$-preference relation $\rho_D$ from a desirability measure $\mathbf{D}$ is easily achieved by means of the pseudoinverse $\div$ of $\oplus$:
$$\rho_D(w|w') = \mathbf{D}(w) \div \mathbf{D}(w').$$

The inverse process of derivation of a unique desirability measure from a preference relation is, in general, not possible. One of several representation theorems of Valverde [Valverde 85]. exploiting in this case the identity
$$\rho(w|w') = \sup_{w'' \text{ in } \mathcal{U}} \left\{\rho(w|w'') \ominus \rho(w'|w'')\right\},$$

assures, however, that there is always a family $\{\mathbf{D}_\alpha\}$ of desirability measures such that
$$\rho(w|w') = \sup_\alpha \left\{\mathbf{D}_\alpha(w) \ominus \mathbf{D}_\alpha(w')\right\}.$$

The above representation has a most natural interpretation as the set of constraints (i.e., desirability measures) that are involved in the generalized order defined by a preference relation, i.e., the criteria that make a solution better than another. As it is often the case with conventional constraints, some of these generalized constraints may never be "active," being, in effect, superseded by more specific restrictions. For this reason, the above decomposition is never unique [Jacas 87]. We may, however, always define a unique "canonical decomposition," which is suggested by the proofs of Valverde's theorems. We will call the family of desirability measures $\{\mathbf{D}_w\}$ defined by
$$\mathbf{D}_w(w') = \rho(w'|w), \qquad \text{for every } w \text{ in } \mathcal{U},$$

the *Valverde representation* of $\rho$.

Note that, although this definition essentially defines a mapping from every possible world $w$ into a desirability measure $\mathbf{D}_w$, the collection of generating functions that is so defined may have a cardinality that is considerably smaller than that of $\mathcal{U}$. The question of whether there exists a unique desirability $\mathbf{D}$ measure that generates $\rho$, i.e., $\rho(w|w') = \mathbf{D}(w) \div \mathbf{D}(w')$, is, in view of the above comments, a matter of rather important practical significance that was studied and solved by Jacas [Jacas 87].

## 4 Combination of Preference Functions

The ability to express any preference function (i.e. relative adequacy of solutions) in terms of a collection of desirability measures (i.e., criteria for adequacy) also suggests a natural algebraic structure for preference relations.

**Definition:** Let $\rho$ and $\rho'$ be two preference relations in the universe of discourse $\mathcal{U}$. Furthermore, let $\{\mathbf{D}_w\}$ and $\{\mathbf{D}'_w\}$ be the Valverde representations of $\rho$ and $\rho'$, respectively. Then the conjunction and disjunction of $\rho$ and $\rho'$ are the preference functions, denoted $\rho \circledast \rho'$ and $\rho \div \rho'$, associated with the generating families $\{\mathbf{D}_w \circledast \mathbf{D}'_w\}$, and $\{\mathbf{D}_w \oplus \mathbf{D}'_w\}$, respectively. Furthermore, the complement of $\rho$ is the preference relation $\sim \rho$ associated with the generating family $\{\sim \mathbf{D}_w\}$. Finally, the implication preference $\rho \rightarrow \rho'$ is the preference relation generated by the family $\{\mathbf{D}_w - \mathbf{D}'_w\}$ of desirability measures.

## 5 Possibility and Necessity

It is often difficult to assess the adequacy of certain solutions (or particular aspects of such solutions), even from the limited perspective provided by specific problem-solving goals. While steering a mobile robot around an obstacle, for example, it is hard to determine if a particular move is preferable to another from the viewpoint of a maneuver to be performed much later at a remote location.

Modal logics [Hughes 68], by introduction of notions of possible and necessary truth, permit to represent states of ignorance about the potential truth of the different statements that are being reasoned about. In the formalism presented in this paper, where restrictive propositions have been generalized as relative measures of solution adequacy, the role of the necessity and possibility operators of modal logic is replaced by lower and upper bounds for measures of desirability and preference so as to generalize the modal implications $\mathbf{N}p{-}p{-}\boldsymbol{\Pi}p$. For example, if we are fully ignorant about the adequacy of $w$ as a solution meeting a constraint represented by the desirability measure $\mathbf{D}$, then we may represent that fact by the bounds $0 \leq \mathbf{D}(w) \leq 1$.

We will say, therefore, that a function $\mathbf{N}_D$ mapping possible worlds $w$ into values between 0 and 1 is a *necessary desirability distribution* for a desirability measure $\mathbf{D}$ if $\mathbf{N}_D(w) \leq \mathbf{D}(w)$ for all $w$ in $\mathcal{U}$. Similarly, we will say that $\boldsymbol{\Pi}_D$ is a *possible desirability distribution* for $\mathbf{D}$ if $\mathbf{D}(w) \leq \boldsymbol{\Pi}_D(w)$ for all $w$ in $\mathcal{U}$.

The following results permit to manipulate necessary and possible desirabilities along lines that generalize similar derivation procedures for conventional modal logic:

(a) If $\mathbf{N}_{\sim D}$ is a necessary desirability for the complement $\sim \mathbf{D}$ of $\mathbf{D}$, then $\sim \mathbf{N}_{\sim D}$ is a possible desirability for $\mathbf{D}$. Similarly, if $\boldsymbol{\Pi}_{\sim D}$ is a possible desirability for the complement $\sim \mathbf{D}$ of $\mathbf{D}$, then $\sim \boldsymbol{\Pi}_{\sim D}$ is a necessary desirability for $\mathbf{D}$. These relations are the generalization of the well-known duality relations $\neg \mathbf{N} \neg p \equiv \boldsymbol{\Pi} p$ and $\neg \boldsymbol{\Pi} \neg p \equiv \mathbf{N} p$.

(b) If $\mathbf{N}_D$ and $\mathbf{N}_{D'}$ are necessary desirability for $\mathbf{D}$ and $\mathbf{D}'$, respectively, then $\mathbf{N}_D \circledast \mathbf{N}_{D'}$ and $\mathbf{N}_D \oplus \mathbf{N}_{D'}$ are necessary desirabilities for $\mathbf{D} \circledast \mathbf{D}'$ and $\mathbf{D} \oplus \mathbf{D}'$, respectively. A similar statement holds for possible desirabilities.

(c) If $\mathbf{N}_D$ is a necessary desirability for $\mathbf{D}$ and if $\boldsymbol{\Pi}_{D'}$ is a possible desirability for $\mathbf{D}'$, then $\mathbf{N}_D \ominus \boldsymbol{\Pi}_{D'}$ is a necessary desirability for $\mathbf{D}'{-}\mathbf{D}$. A dual statement also holds for possible desirabilities.

Bounds, called *necessary* and *possible preference functions*, may also be introduced to represent ignorance about relative preference between solutions. Rules for their manipulation, however, are considerably more complex than those for their desirability counterparts. A rather straightforward consequence, nonetheless, of the definition of preference functions is that if $\mathbf{N}_D$ and $\boldsymbol{\Pi}_D$ are necessary and possibility desirability distributions for a desirability measure $\mathbf{D}$, then the functions defined by the expressions

$$\mathbf{N}_\rho(w|w') = \mathbf{N}_D(w) \ominus \boldsymbol{\Pi}_D(w'),$$

and

$$\boldsymbol{\Pi}_\rho(w|w') = \boldsymbol{\Pi}_D(w) \ominus \mathbf{N}_D(w'),$$

are necessary and possible preferences for $\rho_D(w|w') = \mathbf{D}(w) \ominus \mathbf{D}(w')$.

It should be also clear that necessary and possible preference functions can always be chosen to satisfy the first two properties (generalized nonreflexivity and antisymmetry) of the definition of preference function. Less obvious is the fact that a possible preference function may always be selected to satisfy the third (or transitive) property. Since then such possible preference relation will be itself a preference relation, it may be represented by a family $\mathbf{D}_w$ of desirability measures that is related to the Valverde representation $\mathbf{D}_w$ of $\rho$ by the inequality $\mathbf{D}_w \leq \mathbf{D}_w$.

In closing this section, we may note that, in general, it is more likely that a problem-solver will be interested in issues of desirability of a class of solutions or preference between classes of solutions rather than the the corresponding questions for possible worlds. Unfortunately, it is not possible to characterize such general utilitarian concepts using numeric-valued functions as assessments of the utility of a proposition $p$ as a solution depend on the particular world $w \vdash p$ under consideration. It is possible, however, to define bounds

$$\mathbf{N}_D(p) = \inf_{w \vdash p} \mathbf{D}(w), \quad \text{and} \quad \boldsymbol{\Pi}_D(p) = \inf_{w \vdash p} \mathbf{D}(w),$$

which bound the adequacy of any $p$-world.

Note also that such bounds may be generated from those of possible and necessary desirability distributions for specific solutions. Conversely, values for $\mathbf{N}_D(p)$ and $\boldsymbol{\Pi}_D(p)$ defined for every $p$ in an exhaustive, disjoint, partition $\{p_1, p_2, \ldots p_n\}$ of the universe $\mathcal{U}$ may also be used to define necessary and possible possibility distributions by means of the expressions

$$\mathbf{N}_D(w) = \mathbf{N}_D(p_i), \quad \boldsymbol{\Pi}_D(w') = \boldsymbol{\Pi}_D(p_i), \quad \text{if } w \vdash p_i.$$

The preference of $p$-worlds over $q$-worlds, as measured from the viewpoint of a preference relation $\rho$, may be similarly defined using the expressions

$$\mathbf{N}_\rho(p|q) = \inf_{w \vdash p} \inf_{w' \vdash q} \rho(w|w'),$$

and

$$\boldsymbol{\Pi}_\rho(p|q) = \sup_{w \vdash p} \inf_{w' \vdash q} \rho(w|w').$$

# 6  Preference, Similarity, and Fuzzy Logic

A recent semantic model of the author [Ruspini 91] presented a rationale for the interpretation of the possibilistic structures of fuzzy logic and for its major rule of derivation on the basis of similarity relations between possible worlds. Similarity relations $S$ assign a value $S(w, w')$ between 0 and 1 to every pair of possible worlds $w$ and $w'$ in such a way that

1. $S(w, w) = 1$ for all possible worlds $w$,

2. $S(w, w') = S(w', w)$ for all possible worlds $w$ and $w'$, and

3. $S(w, w') \leq S(w, w'') \circledast S(w'', w')$ for all possible worlds $w$, $w'$ and $w''$, where $\circledast$ is a T-norm.

Two possible worlds $w$ and $w'$ may be considered similar if, from the perspective of all constraints defining a problem, the solutions that they represent have close desirability values. This statement, reflected by the well known relation

$$S(w, w') = \min \left( \sim \rho(w|w'), \sim \rho(w'|w) \right),$$

permits derivation of a similarity relation from a preference relation.

Extensions of the notion of similarity to allow definition of bounds for the resemblance between $p$-worlds and $q$-worlds, called *degree of implication* and *degree of consistence*, which are also the result of applying a similar operation to the corresponding preference bounds $N_\rho(p|q)$ and $\Pi_\rho(p|q)$, play an essential role in the interpretation of the possibility distributions of fuzzy logic.

## Acknowledgements

This work was supported in part by the United States Army Research Office under Contract No. DAAL03-89-K-0156 and in part by a contract with the Laboratory for International Fuzzy Engineering Research. The views, opinions and/or conclusions contained in this note are those of the author and should not be interpreted as representative of the official positions, decisions, or policies, either express or implied, of his sponsors.

The author benefitted from exchanges and conversations with F. Esteva, D. Israel, J. Jacas, J. Lowrance, R. Perrault, E. Trillas, L. Valverde, and L. Zadeh. To all of them, many thanks.